\documentclass[letterpaper, 10 pt, conference]{ieeeconf}  

\IEEEoverridecommandlockouts                              

\usepackage{graphics} 
\usepackage{epsfig} 
\usepackage{mathptmx} 
\usepackage{times} 
\usepackage{amsmath} 
\usepackage{amssymb}  
\usepackage{cite}
\usepackage[dvipsnames]{xcolor}
\usepackage{booktabs}
\usepackage{float}
\usepackage{multirow}
\usepackage{stfloats}
\usepackage{algorithm}
\usepackage{algpseudocode}  
\usepackage{caption}
\usepackage{cuted}
\usepackage{newtxtext, newtxmath}
\usepackage{hyperref}
\usepackage{balance}

\newcommand{\systemname}{\text{SARO}}
\newcommand{\policyname}{\text{PAS}}

\title{\LARGE \bf
SARO: Space-Aware Robot System for Terrain Crossing \\ \textit{via} Vision-Language Model 
}

\author{Shaoting Zhu*$^{1,2}$, Derun Li*$^{1,3}$, Linzhan Mou$^{1,5}$, Yong Liu$^{4}$, Ningyi Xu$^{3}$, Hang Zhao†$^{1,2}$
\thanks{$^{1}$Shanghai Qi Zhi Institute, Shanghai, China}%
\thanks{$^{2}$IIIS, Tsinghua University, Beijing, China }%
\thanks{$^{3}$SEIEE, Shanghai Jiao Tong University, Shanghai, China }%
\thanks{$^{4}$CSE, Zhejiang University, Hangzhou, China }%
\thanks{$^{5}$GRASP Lab, University of Pennsylvania, Philadelphia, PA, USA}%
\thanks{* These authors contributed equally to this work.}
\thanks{† Corresponding author. E-mail: \texttt{hangzhao@mail.tsinghua.edu.cn}}%
}

\begin{document}
\maketitle
\thispagestyle{empty}
\pagestyle{empty}

\begin{strip}
\begin{minipage}{\textwidth}
\centering
\vspace{-60pt}
\includegraphics[width=\textwidth]{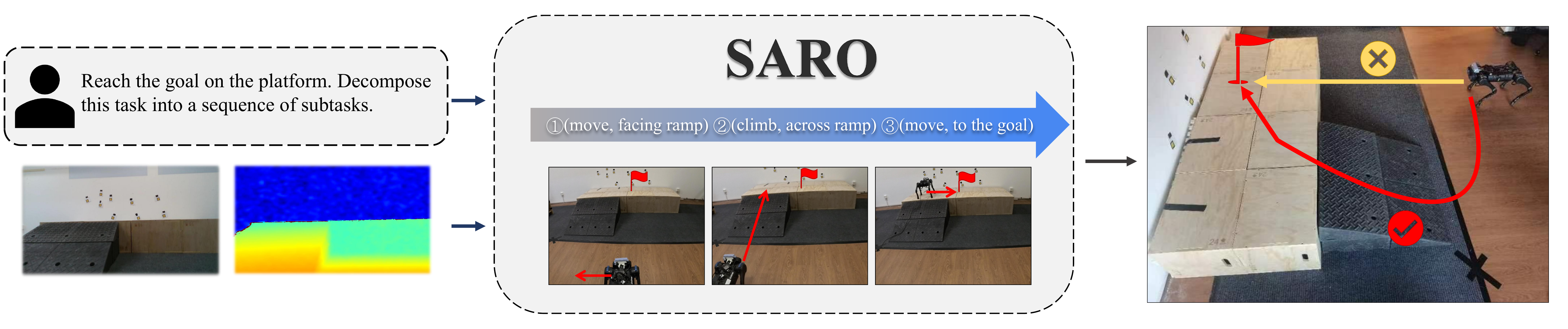}
\captionof{figure}{\systemname\ achieves space-aware navigation capability for 3D terrain crossing. Our system utilizes the reasoning and motion planning ability of the VLM model, with a design of task decomposition and a closed-loop sub-task execution module. While traditional navigation approaches (\textcolor{Tan}{yellow}) fail in reasoning 3D environment, our system (\textcolor{red}{red}) guides the quadruped robot to cross the accessible intermediation towards the goal.}
\vspace{-3mm}
\label{fig:figure1}
\end{minipage}
\end{strip}

\begin{abstract}
    The application of vision-language models (VLMs) has achieved impressive success in various robotics tasks. However, there are few explorations for these foundation models used in quadruped robot navigation through terrains in 3D environments. In this work, we introduce \systemname~(\underline{S}pace-\underline{A}ware \underline{Ro}bot System for Terrain Crossing), an innovative system composed of a high-level reasoning module, a closed-loop sub-task execution module, and a low-level control policy. It enables the robot to navigate across 3D terrains and reach the goal position. For high-level reasoning and execution, we propose a novel algorithmic system taking advantage of a VLM, with a design of task decomposition and a closed-loop sub-task execution mechanism. For low-level locomotion control, we utilize the \underline{P}robability \underline{A}nnealing \underline{S}election (\policyname) method to effectively train a control policy by reinforcement learning. Numerous experiments show that our whole system can accurately and robustly navigate across several 3D terrains, and its generalization ability ensures the applications in diverse indoor and outdoor scenarios and terrains.
    \textit{Appendix} and \textit{Videos} can be found in project page: \href{https://saro-vlm.github.io/}{https://saro-vlm.github.io/}.
\end{abstract}

\section{INTRODUCTION}
\label{sec:introduction}
	
    The athletic intelligence of animals is concentrated in their understanding of complex wild environments and their ability to reach invisible destinations. This significantly challenges the capacity for 3D scene understanding and traversability across various terrains. It should be considered as the potential advantage for quadruped robot agents. Although much progress has been made in specific locomotion skills \cite{kumar2021rma, margolis2023walk}, the autonomy of robots should be improved at the system level.
    

    Vision-language models (VLMs) have shown advancements in common sense reasoning and remarkable generalization in vision tasks, significantly boosting the progress of robotic learning\cite{li2023manipllm, liu2024moka, huang2024copa, dream2real, myers2024policy, huang2024rekep, chen2024commonsense}. However, VLMs suffer from the limitations of training data perspectives and the lack of a memory information bank, which is believed to curtail their usage in robot navigation tasks. Our motivation originates from this important question: \textit{How can we design a system to fully activate the potential of VLMs' visual common sense on robots to enable them to observe and understand and travel in the 3D world?} In this work, we design a system called \systemname, composed of a high-level reasoning module, a close-loop sub-task execute module, and a low-level control policy. The system enhances the 3D reasoning, motion planning, and locomotion ability of the robots. Our design uses zero-shot VLM common sense reasoning to overcome the lack of training data and utilizes the closed-loop sub-task execution to invest a memory-free mechanism to transfer agents stage by stage in the navigation process.

     Besides, the low-level locomotion control policy needs to be adaptable to different terrains and robust against diverse environments. Traditional control methods such as SLIP\cite{poulakakis2006stability}, VMC\cite{pratt1997virtual}, MPC\cite{bledt2018cheetah,di2018dynamic} can handle some specific terrain tasks, but they have poor robustness against complex real-world situations. Adaptation learning \cite{kumar2021rma} and teacher-student framework\cite{lee2020learning} are employed to address the transferring from simulation to real world. However, they are prone to significant performance degradation when deployed in the real world. In our work, we propose a novel method called \textbf{P}robability \textbf{A}nnealing \textbf{S}election (\policyname) to solve the Oracle policy transfer problem caused by mimic learning. It comprehensively learns the ability to cross various types of real-world 3D terrains.

    We test our method across several different categories of terrains and intermediations. In addition, we showcase the capacity of our low-level locomotion policy in extra experiments. Since the restricted view range of the front view camera input, we define the task as goal-tracking with a simple setting of only one intermediation chosen from ``stair'', ``ramp'', ``gap'', and ``door'' for each task in experiments. Our experiment results demonstrate the generalization and common-sense visual motion reasoning abilities of \systemname\ in quadruped robot 3D navigation. Our method exhibits generalization ability and robustness in real-world scenarios, as demonstrated in our real-world videos. 
    
    In summary, our contributions are as follows:

    \textbf{1)} We innovatively present \systemname, a system that is composed of a high-level reasoning module, a closed-loop sub-task execution module, and a robust low-level control policy. This novel system enhances ego-view 3D navigation for quadruped robots.
    
    \textbf{2)} We introduce a novel reinforcement learning-based locomotion control policy \textbf{P}robability \textbf{A}nnealing \textbf{S}election (\policyname) to overcome various 3D terrain challenges.
    
    \textbf{3)} We conduct experiments across extensive terrains. The results demonstrate our system's ability to complete our specially defined goal-tracking task across 3D terrains.

\section{RELATED WORKS}
\begin{figure*}[tbp]
    \centering
    \includegraphics[width=\linewidth]{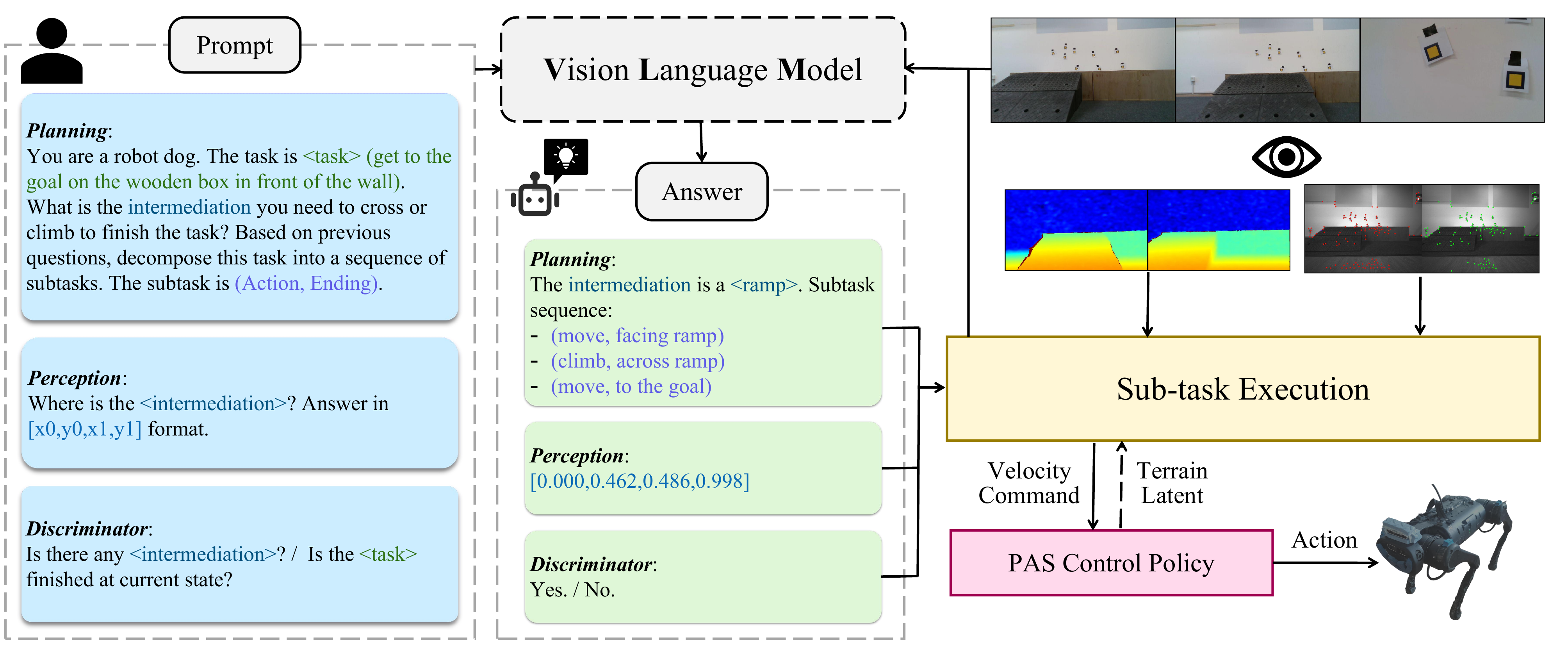}
    \caption{Overview of the SARO system. The robot needs to complete a goal-tracking task while autonomously navigating through a 3D terrain. The pretrained vision-language foundation model (VLM) takes as input RGB images and prompts querying the 3D environment perception to decompose the task into sub-tasks. After that, the system executes these sub-tasks in a closed-loop taking advantage of VLM discriminator double-check. The well-designed sub-task execution module connects the high-level VLM and the low-level control policy and receives depth image and stereo image information to help localization.}
    \label{fig:system}
    \vspace{-5mm}
\end{figure*}

\label{sec:related_work}

\subsection{Foundation Models in Robotics}
    In recent years, foundation models \cite{devlin2018bert, chowdhery2023palm, achiam2023gpt, li2022blip, liu2023llava}, including large language models (LLMs) and vision-language models (VLMs), have achieved significant advancements. Furthermore, some of these foundation models have been adapted to the domain of robotics \cite{hu2023toward,brohan2023can, huang2022inner, chen2023open, huang2022language, xu2023creative, ouyang2024long, liang2023code, singh2023progprompt, ma2023eureka, alakuijala2023learning, mahmoudieh2022zero, ma2023liv}. Several works \cite{stone2023open, chen2023open, gadre2023cows, yokoyama2023vlfm} have utilized open-vocabulary pretrained models for robotic tasks, Others \cite{hu2023look, li2023manipllm, liu2024moka, huang2024copa, tian2024drivevlm} have employed powerful VLMs, such as GPT-4V, for robotics tasks. Moreover, some researchers have attempted to apply foundation models to quadruped robots. Saytap \cite{tang2023saytap} uses large language models (LLMs) to translate natural language commands into foot contact patterns for quadrupedal robots. ViNT \cite{shah2023vint} trains a universal policy from a large-scale visual navigation dataset using the Transformer \cite{vaswani2017attention}. CognitiveDog \cite{lykov2024cognitivedog} integrates a Large Multi-modal Model (LMM) with a quadruped robot. GeRM \cite{song2024germ} trains a generalist model for quadruped robots in vision-language tasks. QuadrupedGPT\cite{wang2024quadrupedgpt} and Commonsense\cite{chen2024commonsense} utilize large models for movement in simple scenes. Nevertheless, all these methods are only suitable for tasks on planar surfaces and do not fully utilize the 3D terrain capabilities of quadruped robots.
    
\subsection{Locomotion Control of Quadruped Robots}
    Traditional locomotion control methods for quadruped robots \cite{poulakakis2006stability, pratt1997virtual, mistry2010inverse, sleiman2021unified, grandia2023perceptive} is one way for locomotion control, but they often suffer the unstable problem in real-world deployment. Reinforcement learning has shown remarkable capabilities in recent years\cite{hwangbo2019learning, lee2020learning, kumar2021rma}. They utilize the privileged training paradigm to train quadruped robots without extra sensors. Also, some work \cite{miki2022learning, agarwal2023legged, zhuang2023robot, cheng2023extreme, luo2024pie} integrate proprioceptive and exteroceptive states to achieve agile locomotion.
    Mimic learning is frequently used in previous works. Methods of adaptation learning \cite{kumar2021rma, agrawal2022vision, margolis2023walk} and teacher-student framework learning \cite{lee2020learning, margolis2024rapid, wu2023learning} are used to solve the sim-to-real transfer problem, but they suffer from high-performance reduction during real deployment. Meanwhile, some other works propose innovative methods to improve locomotion efficiency. DayDreamer \cite{wu2023daydreamer} learns a ``world model'' to synthesize infinite interactions, while DreamWaQ \cite{nahrendra2023dreamwaq} implicitly infers terrain properties and adapts its gait accordingly by learning a VAE model. Traditional control methods are combined with deep reinforcement learning \cite{long2023hybrid, long2024learning, chen2024slr} to accelerate training speed, but they do not fully utilize privileged information in simulation with only one-stage training.
    In our work, we propose a novel method to solve the Oracle policy transfer problem without mimic learning, while fully taking advantage of privileged information in simulation with two-stage training.

\section{METHOD}

\label{sec:method}

\subsection{Task Definition}
    The task is defined as a goal-tracking task for the quadruped robot to autonomously navigate through a 3D environment with various terrains. A terrain $\mathcal{T}$ is composed of two platforms $\mathcal{P}_1$, $\mathcal{P}_2$ and one intermediation $\mathcal{I}$ connecting two platforms in 3D space. In the beginning, the robot is located on $\mathcal{P}_1$, and the task is to reach a specified goal $\mathcal{G}$ on $\mathcal{P}_2$ defined as $(x, y, z, yaw)$ relative to the robot's starting position, combined with a language description $\mathcal{L}$ of the goal. The robot needs to cross intermediation $\mathcal{I}$ to reach the goal on platform $\mathcal{P}_2$. The 3D intermediation in this work includes ``stairs'', ``ramps'', ``gaps'', and ``doors'', which the robot does not take extreme action to come across. The quadruped robot can only access the sensors onboard, including proprioception, ego-view RGB image, and depth image. In summary, the tasks can be represented as:
    \begin{align*}
        \textit{Find the way and navigate through:}\ \{\mathcal{P}_1 \rightarrow \mathcal{I} \rightarrow \mathcal{P}_2\} \\ \textit{under the condition of} \ \mathcal{G} \ \textit{and} \ \mathcal{L}
    \end{align*}
    As an example, in Fig.~1, the robot begins on the floor ($\mathcal{P}_1$) in front of a higher platform ($\mathcal{P}_2$) and needs to get the goal $\mathcal{G}$ located on the platform. The goal point is ``$(3.0m, 0.0m, 0.4m, 0.0rad)$''. The language instruction $\mathcal{L}$ is ``getting to the goal on the wooden box in front of the wall''. We list all of our experiment scenario details in Section ~\ref{sec:result}.
    
\subsection{High-level Reasoning and Task Execution}

\textbf{Task Decomposition: }
Our system works as a state machine and decomposes the multi-step navigation into a sub-task sequence composed of movement actions and the ending point by prompting the VLM. The prompt is based on the task that defines the robot's ability and instructs it to cross the terrain. As illustrated in Fig.~\ref{fig:system}, we use the pre-trained VLM to perform zero-shot inference on ego-view image inputs. It first recognizes the intermediation $\mathcal{I}$ associated with the task instruction $\mathcal{L}$. After that, the VLM can further generate the decomposed sub-task sequence. The sub-task is defined as an \textit{(Action, Ending)} pair. \textit{Action} is one of [``move'', ``climb'']. \textit{Ending} is one of [``facing intermediation'', ``across intermediation'', and ``to the goal'']. The full prompts and examples are shown in the \textit{Appendix A.1}.
    
\begin{figure}[h]
    \centering
    \includegraphics[width=0.98\linewidth]{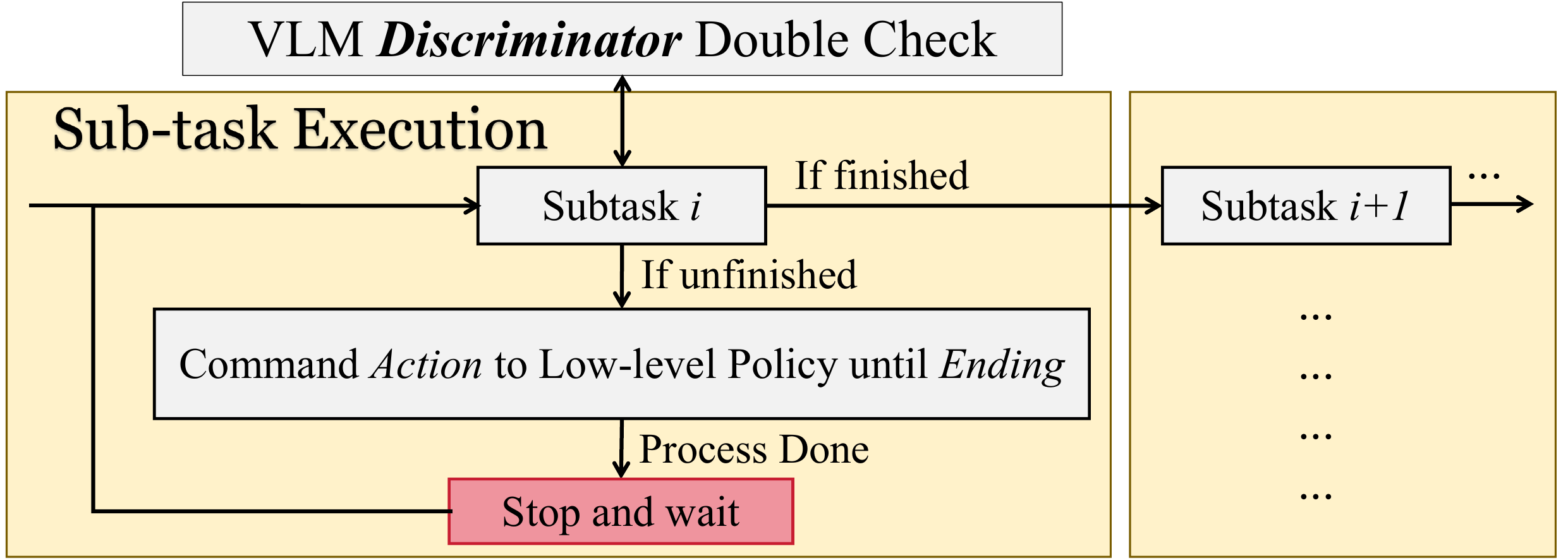}
    \caption{After task decomposition, the system executes the sub-tasks one by one. The double-check closed-loop module improves the robustness of the system.}
    \vspace{-5mm}
    \label{fig:ring}
\end{figure}

\textbf{Sub-task Execution: }We extensively explore the perception abilities of VLM to aid in fine-grained trajectory guidance and judgment of sub-task states. As shown in Fig.~\ref{fig:ring}, for each sub-task, the VLM discriminator first judges whether the sub-task is finished based on the \textit{Ending}. If it is unfinished, a velocity command will be sent to low-level policy based on the \textit{Action} and VLM's language instruction. The predefined execution workflow determines how to complete this \textit{Action} until the \textit{Ending} point, which is expanded in detail in the \textit{Appendix A.2}. Both sub-task execution workflow and VLM discriminator can judge the \textit{Ending} point, but the system will only be permitted to conduct the next sub-task if the VLM discriminator outputs the ending signal, which we call double-check. It is worth noting that one sub-task may take several processes to execute because the low-level workflow may not predict an accurate \textit{Ending} point. For example, if the intermediation is not completely within the field of view, several adjustments are required to do center alignment. This closed-loop module and double-check mechanism fully leverage sensor input from the quadruped robot, enhancing the robustness and safety of our system.

\subsection{Low-level Locomotion Control Policy}
\label{subsec:locomotion}

\begin{figure*}[tbp]
    \centering
    \includegraphics[width=0.8\linewidth]{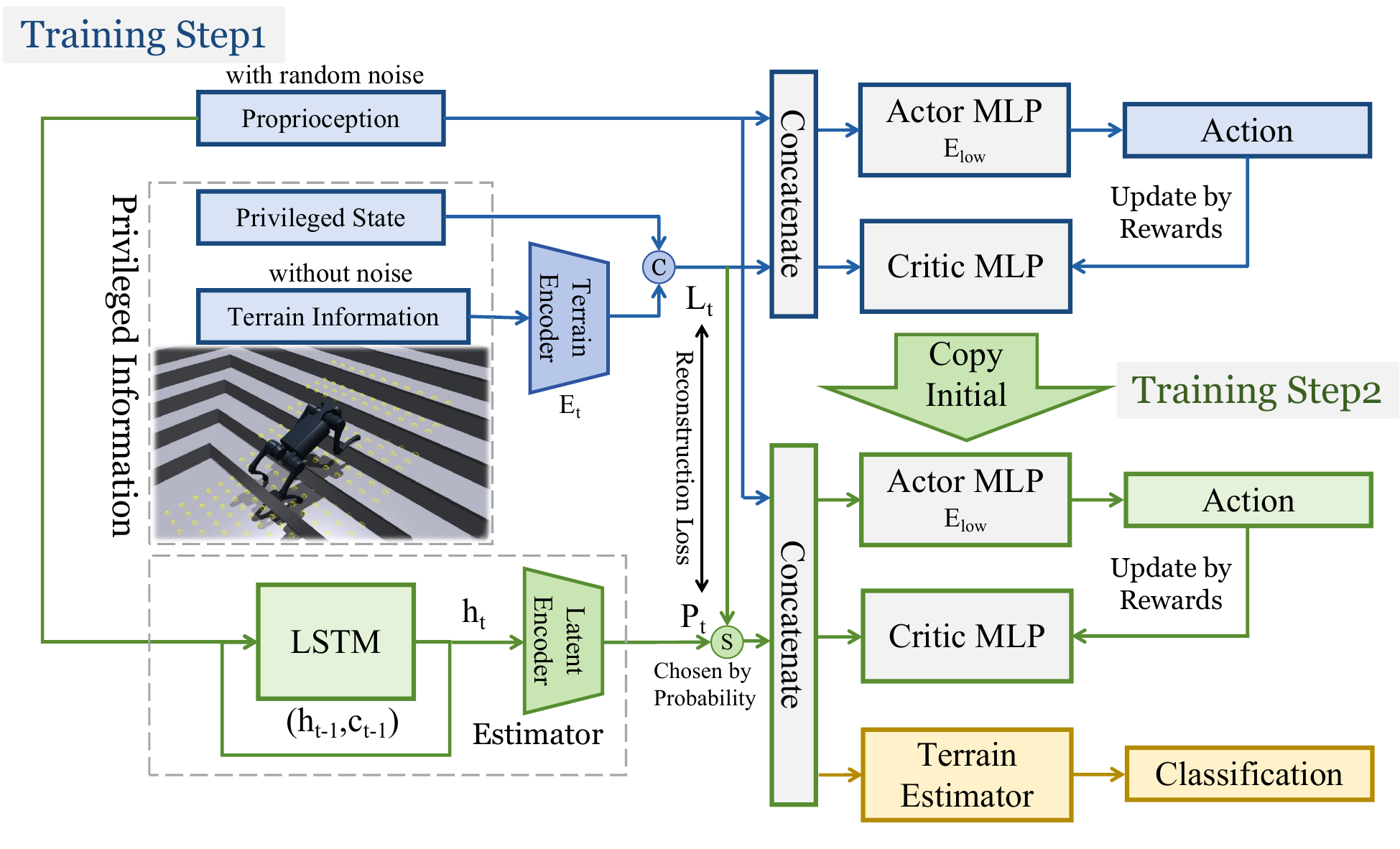}
    \caption{Overview of the low-level locomotion control policy. In the first training step, we train an oracle policy using proprioception $\boldsymbol{p}_t\in\mathbb{R}^{45}$, privileged state $\boldsymbol{s}_t\in\mathbb{R}^{4}$, and terrain information $\boldsymbol{t}_t\in\mathbb{R}^{187}$. In the second training step, we use the probability annealing selection (\policyname) method to train the final actor network, which only uses proprioception as input. After the policy training process is finished, we exclusively train a terrain estimator to classify whether the robot is on the plane or is climbing the intermediation.}
    \label{fig:pipeline}
    \vspace{-6mm}
\end{figure*}
    
We enable the quadruped robot to track expected linear and angular velocities over various terrains with only one policy. The reinforcement learning policy only trains on proprioception, without any additional external perception such as depth camera and Lidar.

\textbf{Oracle Policy Training: }In the first training step, we train an oracle policy. All information serves as the policy observation, including proprioception $\boldsymbol{p}_t\in\mathbb{R}^{45}$, privileged state $\boldsymbol{s}_t\in\mathbb{R}^{4}$, and terrain information $\boldsymbol{t}_t\in\mathbb{R}^{187}$. The terrain information is first encoded by terrain encoder $E_{t}$ into terrain latent $\boldsymbol{t}_{l_t}\in\mathbb{R}^{32}$. Then it is concatenated with $\boldsymbol{s}_t\in\mathbb{R}^{4}$ into full latent state $\boldsymbol{l}_t\in\mathbb{R}^{36}$. This accelerates the training convergence and enhances the training stability. The input of the low-level network $E_{low}$ $\boldsymbol{o}\in\mathbb{R}^{81}$ is composed of proprioception $\boldsymbol{p}_t\in\mathbb{R}^{45}$ and $\boldsymbol{l}_t\in\mathbb{R}^{36}$. The actor outputs the desired joint positions $\boldsymbol{a}_t\in\mathbb{R}^{12}$. The critic is also an MLP, but it directly uses the concatenation of three different types of information $\boldsymbol{o}_c\in\mathbb{R}^{236}$. Due to the use of privileged information, the quadruped robot can quickly and effectively learn locomotion skills on various terrains. While training the Oracle policy, we also train a state estimator network $E_{e}$ concurrently. Mean Squared Error (MSE) loss is used to reconstruct the latent of privileged information $\boldsymbol{l}_t\in\mathbb{R}^{36}$.

\textbf{Partial Observation Policy Training: }In the second training step,
 we train the estimator network and the low-level MLP network jointly using the \textbf{P}robability \textbf{A}nnealing \textbf{S}election (\policyname) method. The input of the estimator is proprioception $\boldsymbol{p}_t\in\mathbb{R}^{45}$. After passing through the LSTM network, the output is then fed into the MLP encoder to get the latent state prediction $\boldsymbol{p}_t\in\mathbb{R}^{36}$. At the beginning of the second training step, the estimator and the low-level MLP are copied to initialize from the first training step. Then, the loss of reinforcement learning is utilized to simultaneously optimize both the estimator and the low-level MLP. Then, the \policyname\ method is used. Specifically, at the beginning, the latent state vectors input to the lower-level MLP are mostly real values $\boldsymbol{l}_t$, with a small portion of predicted values $\boldsymbol{p}_t$. As the training iterations increase, the probability of selecting real values gradually decreases, and the probability of selecting predicted values gradually increases. Ultimately, only predicted values are used. Specifically, 

\begin{align}
\vspace{-8mm}
\boldsymbol{p}_t &= E_{e}(\boldsymbol{o}_t^p),\\
\boldsymbol{i}_t &= \text{Probability Selection~}(\boldsymbol{P}_t, \boldsymbol{p}_t, \boldsymbol{l}_t),\\
\boldsymbol{a}_t &= E_{low}(\boldsymbol{i}_t, \boldsymbol{o}),\\
\text{Probability~} \boldsymbol{P}_t &= \boldsymbol{\alpha}^{iteration}.
\vspace{-4mm}
\end{align}

To ensure consistency in the same continuous action, the probability selection is based on the number of robots. Our training method ensures the stability of the training process and enhances the performance of the final policy. Training details including problem definition, reward function, termination conditions, terrain curriculum, dynamic randomization, network architecture, and hyper-parameters are shown in the \textit{Appendix B}.


\section{EXPERIMENTS}
\label{sec:result}

\subsection{Experiment Setup}

We deploy our method on Unitree A1 quadruped robot with NVIDIA Jetson Xavier NX as the onboard computer. Also, we use a laptop and a GPU server as the computation platform. The low-level locomotion control policy runs on onboard Xavier NX at 50 Hz. 

\begin{figure}[htbp]
    \centering
    \includegraphics[width=\linewidth]{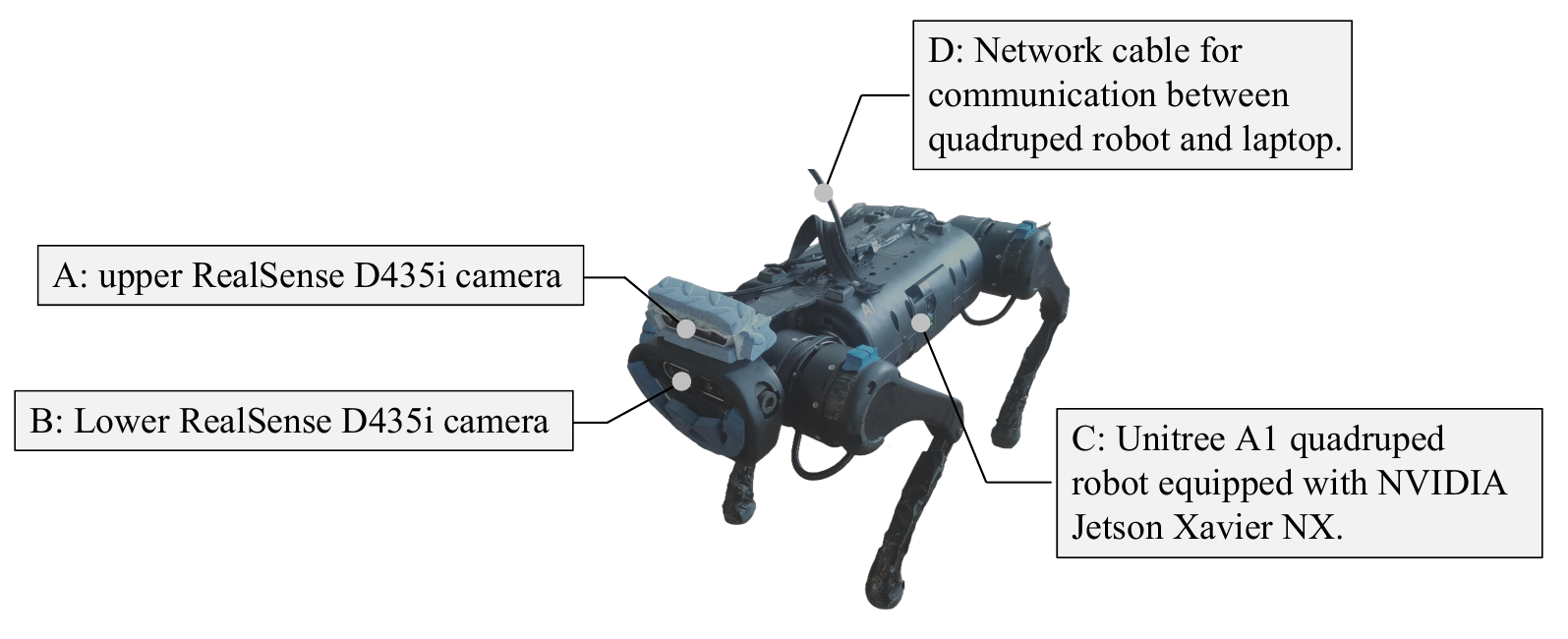}
    \caption{Robot setup for experiment.}
    \vspace{-4mm}
    \label{fig:config}
\end{figure}

\begin{figure}[htbp]
    \centering
    \includegraphics[width=1.0\linewidth]{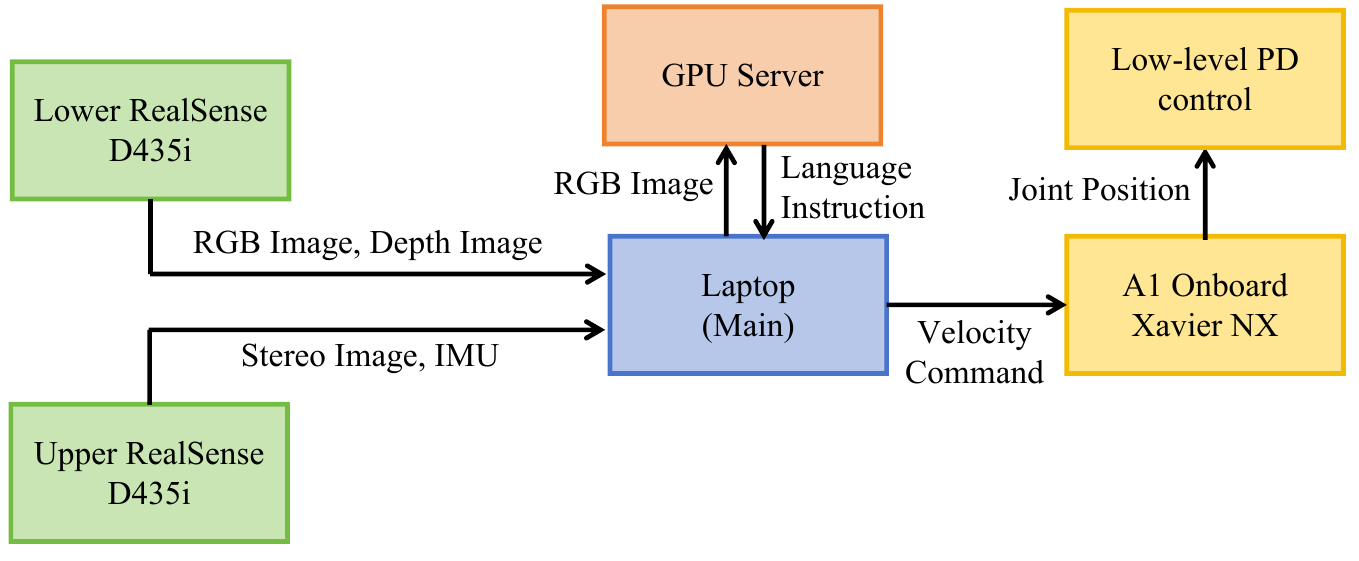}
    \caption{Communication system setup of robot deployment.}
    \label{fig:robot_system}
    \vspace{-4mm}
\end{figure}

We deploy two RealSense D435i cameras on the quadruped robot, the upper one for VIO to get robot odometry, and the lower one for high-level reasoning. The upper one takes stereo images at a resolution of 640$\times$480. The lower one takes an aligned RGB-D image at a resolution of 848$\times$480. We use LLaVA-34B\cite{liu2024llavanext} as both vision language model and vision language model discriminator, VINS-Fusion\cite{qin2019a} as the VIO algorithm. Detailed robot setup for experiment is shown in Fig.~\ref{fig:config}. Based on ROS, we set up our communication system as shown in Fig.~\ref{fig:robot_system}. We use a Ubuntu 20.04 laptop as the main computer, a GPU server with 8 NVIDIA RTX 3090 as the side computer, and NVIDIA Jetson Xavier NX as the onboard computer. The messages from cameras are directly sent to the laptop. We use the laptop to run the SLAM program and the system's main program. The GPU server runs the LLaVA program and communicates RGB images and language instructions with the laptop. The onboard Xavier NX receives the velocity commands by the main program from the laptop through ROS messages and runs the low-level control policy to predict desired joint positions for PD control. Note that we paste some rectangle decorations on the white wall of the laboratory, which is only to provide feature points for the VIO algorithm and improve its stability.

\subsection{Results of High-level Reasoning}

\begin{figure*}[htbp]
    \centering
    \vspace{2mm}
    \includegraphics[width=\linewidth]{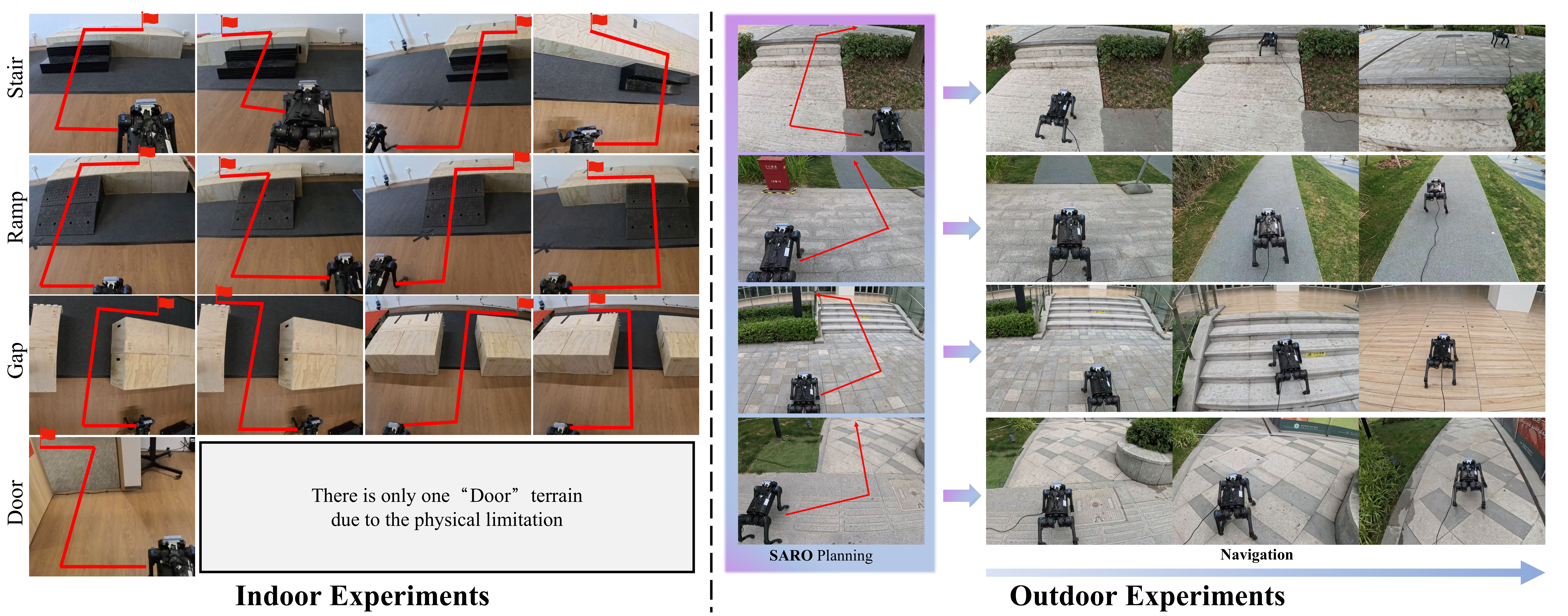}
    \caption{Indoor and outdoor 3D navigation experiment results on diverse terrains.}
    \label{fig:exp}
    \vspace{-2mm}
\end{figure*}

\textbf{Indoor Experiments: }To evaluate our crossing system, experiments are conducted based on versatile routes in the real world, shown in Fig~\ref{fig:exp}. We test the robustness on different terrains including stairs, ramps, gaps, and doors. For every terrain, the goals are at various directions. 20 trials are conducted for each situation and we record the success rate of the whole process, only crossing terrains and stable localization, which means that the localization module always works accurately, respectively as the evaluating metrics. We compare our results with three baselines: naive LSTM network, ViNT\cite{shah2023vint}, and NoMaD\cite{sridhar2023nomad}. We record a real-world dataset containing 50 trajectories including routes on the plain, across stairs, and through ramps, with a frame rate of 15Hz and length of about 10 seconds. A naive baseline is implemented, composed of a CNN image encoder, an LSTM backbone, and an MLP decoder. We train this baseline on the dataset for 500 epochs using MSE loss. We also collect the topological maps and the goal images in the real world for ViNT\cite{shah2023vint} to simulate our goal-pursing task. We implement NoMaD\cite{sridhar2023nomad} for terrain crossing tasks using its exploration function. The results are shown in Table~\ref{tab:highlevel}.

\begin{table*}[htbp]
    \centering
    \fontsize{10}{12}\selectfont
    \vspace{4mm}
    \caption{\label{tab:highlevel}The success rate of high-level navigation in versatile real-world experiments. Gray ones indicate the success rates of only crossing terrain sub-tasks (no need to get stop at the goal).}
    \vspace{2mm}
    \begin{tabular}{c|ccccccc}
        \toprule
        \textbf{Intermediation} & \textbf{Overall} & \textbf{Stable Loc} & \textbf{w/o Closed-Loop} & \textbf{NoMaD} & \textbf{LSTM} & \textcolor{gray}{\textbf{Across Terrains}} & \textcolor{gray}{\textbf{ViNT}} \\ 
        \midrule
        Stair & 60\% & \textbf{88\%} & 10\% & 0\% & 0\% & \textcolor{gray}{70\%} & \textcolor{gray}{0\%}\\
        Ramp & 25\% & \textbf{67\%} & 10\% & 0\% & 0\% & \textcolor{gray}{50\%} & \textcolor{gray}{0\%}\\
        Gap & 45\% & \textbf{94\%} & 20\% & 20\% & 0\% & \textcolor{gray}{80\%} & \textcolor{gray}{0\%}\\
        Door & 30\% & \textbf{63\%} & 15\% & 70\% & 0\% & \textcolor{gray}{50\%} & \textcolor{gray}{0\%}\\
        \bottomrule
    \end{tabular}
    \vspace{-3mm}
\end{table*}

We observe relatively good results for versatile intermediations, especially for stairs, which demonstrates the effectiveness and robustness of our crossing system. The task definition is much harder than NoMaD and ViNT since the starting point is closer to the intermediations and it requires faster planning and adjustment to achieve the correct position and direction, which blocks the baselines from reacting well. Also, all three baselines lack the 3D reasoning and planning capability as \systemname\ holds. It is worth noting that our overall success rate relies heavily on the accuracy of our localization module. Our ablations on crossing intermediation and the ideal stable localization module show a large margin of improvement in all of the situations. This encourages our valuable system design and great cooperation of generalizable control policy and other modules. Meanwhile, we point out that most of the localization errors occur when the input images are blurred due to the high-dynamic motion, especially for the ramps which makes the quadruped robot lean upwards.

VLM actively participates in every step of the 3D navigation task and demonstrates its strong power of common sense reasoning and motion estimation. While several challenges, such as transforming ego-view 2D image information into 3D environment interactions, arise in the application of VLM, we manage to bridge the gap by designing sub-task execution modules that leverage rich information from other robot sensors. The entire system is capable of generalizing to multiple 3D terrains and diverse environments. Additionally, it can be robustly embedded in real-world quadruped robots.

\textbf{Outdoor Experiments: }Additionally, we test our system outdoors to show the generalizing ability. As shown in Fig~\ref{fig:exp}, we spot that our framework can be easily extended to the wild environment and the versatile 3D terrain conditions can be covered by our robust perception, planning, and control pipeline. More details can be referred to in our appended demo videos.

\subsection{Results of Low-level Locomotion}

To further evaluate our proposed \policyname\ method, we conduct extra experiments on the lower-level locomotion control. We select a variety of challenging terrains, including uneven ground, stairs, ramps, and unseen terrains, to test the success rate and speed tracking ratio. Both metrics are the higher the better.


\textbf{Simulation Results: }For success rate, we conduct 4,096 independent experiments for each type of terrain. If the robot reaches the edge of the terrain or is alive for over 20 seconds, it is marked as a success. The velocity tracking ratio shows the performance for tracking velocity commands, and we also conducted 4,096 experiments for each terrain. First, we compare our method with several previous baselines using only proprioception. The details of methods used for comparison can be found in \textit{Appendix C.2}. Results show that our method significantly outperforms previous methods. Both RMA (latent space) and the ``teacher-student'' framework (action space) only utilize imitation learning in the second step of training without reward feedback, and they perform worse than those of pure blind training and concurrent training. Pure blind training, which does not use privileged information for training, performs relatively well on flat ground and simpler terrains. But its performance drops rapidly once the terrain becomes more complex. Concurrent training only uses partial perceptual information, so it is difficult to estimate an accurate latent state.

\begin{table}[htbp]
    \centering
    \caption{Comparison experiment results in simulation}
    \setlength\tabcolsep{2pt}
    \fontsize{8}{9}\selectfont
    \begin{tabular}{c|ccccc}
        \toprule
        \textbf{Metric} & \textbf{\policyname (ours)} & \textbf{RMA}\cite{kumar2021rma} & \textbf{IL}\cite{wu2023learning}  & \textbf{Blind} & \textbf{Concurrent}\cite{ji2022concurrent} \\ \midrule
        SR (Avg)  & \textbf{85.31\%}   & 49.30\%   & 58.05\%  & 77.34\% & 70.40\% \\
        L vel (Avg)       & \textbf{0.8790}   & 0.6848   & 0.8029  & 0.8555 & 0.8330 \\
        A vel (Avg)      & \textbf{0.7815}   & 0.5895  & 0.7370  & 0.7627 & 0.7349 \\
        \bottomrule
    \end{tabular}
\end{table}

We also conduct ablation experiments for different annealing settings. Results show that the exponential annealing with the base of 0.9998 performs the best in most scenarios. Exponential annealing with an annealing probability of 0.9995 may be too rapid. No annealing results in the robot being unable to adapt at the start of learning, necessitating a period of re-exploration, during which a significant amount of the oracle policy's capabilities are lost. Furthermore, we find that although cosine annealing intuitively follows a slow-fast pattern, its performance is the worst, especially in terms of angular velocity tracking ratio, which may be due to a rapid collapse of the network at a certain point.

\begin{table}[htbp]
    \centering
    \caption{Ablation experiment results in simulation}
    \setlength\tabcolsep{2pt}
    \fontsize{8}{9}\selectfont
    \begin{tabular}{c|ccccc}
        \toprule
        \textbf{Metric} & \textbf{Exp 0.9998} & \textbf{Exp 0.9995} & \textbf{No anneal}  & \textbf{Cosine} & \textbf{Linear} \\ \midrule
        SR (Avg) & \textbf{85.31\%}   & 83.60\%   & 83.33\%  & 83.26\% & 84.00\% \\
        L vel (Avg)      & \textbf{0.8790}   & 0.8710   & 0.8687 & 0.8730 & 0.8715 \\
        A vel (Avg)    & \textbf{0.7815}   & 0.7741  & 0.7733  & 0.7466 & 0.7660 \\
        \bottomrule
    \end{tabular}
\end{table}

\textbf{Real-world Results: }We conduct experiments across a series of terrains. For experiments on each type of terrain, we continuously conduct 20 trials. For each trial, if the robot could start from the beginning, and reach the end without falling or getting stuck, it is considered as a success. As the results shown in Table~\ref{tab:real-exp}, our robot can pass through terrains blindly and is highly competitive with previous methods.

\begin{table}[htbp]
    \centering
    \caption{\label{tab:real-exp}Success rate results in real-world experiments}
    \begin{tabular}{c|cccc}
        \toprule
        \textbf{Terrain type}  & \textbf{\policyname (ours)}  & \textbf{RMA} & \textbf{IL}  & \textbf{Built-in MPC} \\ \midrule
        Stair    & \textbf{100\%}   & 75\%   & 80\%  & 0\% \\
        Ramp     & \textbf{90\%}   & 70\%   & 80\%  & 55\% \\
        Rubble    & \textbf{95\%}   & 70\%   & 75\%  & 0\% \\
        Grassland   & \textbf{100\%}    & 80\%   & 95\%  & 60\%    \\
        Unseen obstacle  & \textbf{95\%}    & 65\%   & 80\%  & 0\%    \\
        \bottomrule
    \end{tabular}
    \vspace{-2mm}
\end{table}

\section{CONCLUSION}
\label{sec:conclusion}

We present a space-aware robot system (\systemname) for vision navigation in 3D environments. The high-level module utilizes task decomposition and closed-loop sub-task execution module to boost the 3D scene understanding and motion planning. The low-level control policy \policyname\ is designed as a novel reinforcement learning method that efficiently learns a partial policy from the oracle policy and facilitates the quadruped robot crossing versatile 3D terrains. Our extensive experiments in both simulator and real-world demonstrate the effectiveness and robustness of the whole system as well as the locomotion control policy. 

For limitation, due to the high-frequency vibrations of the quadruped robot bringing errors to IMU and blurring the image, the current commonly used SLAM method is not as stable as we expect, and it damages the reliability of our whole system. In addition, we only deploy our system on short-term simple tasks, which is limited by ego-view perception and lack of memory. In the future, we believe that better perception and localization approaches can benefit our pipeline. Topological maps or semantic maps can be integrated with VLM to help investigate more complex tasks.




\clearpage
\newpage
\section*{APPENDIX}

In \textit{Appendix}, we introduce technical details of our system implementation, training and experiments. The \textit{Appendix} includes three parts for each, as listed below:

\begin{itemize}
    \item \textbf{A. Details of High-level Reasoning and Task Execution}: the implementation details of our high-level system, including the VLM prompts, sub-task set and the deployed algorithm.
    \item \textbf{B. Training Details of Low-level Locomotion Control Policy}: the training details of our low-level locomotion control policy, including the problem definition, reward function, training strategy and parameters.
    \item \textbf{C. Extra Experiments Details of the Low-level Locomotion}: the supplementary experiment of our low-level locomotion control policy, including the metrics, comparison with baselines and ablation studies.
\end{itemize}

\subsection{Details of High-level Reasoning and Task Execution}

\subsubsection{VLM Details and Prompts}
\label{subsec:VLM}

We use pretrained LLaVA-34B for VLM inference. The VLM takes RGB images at a resolution of 848$\times$480 and instruction prompts as input. The prompts given for different stages are listed below:
\begin{enumerate}
    \item \textbf{Planning} ``Ignore anything on the wall. You are a robot dog. The intermediation may be a stair, a ramp, a gap, or a door frame. The task is \textit{task}. First answer the question: 1. What is the only intermediation you need to cross or climb to finish the task? Based on previous questions, decompose this task into a sequence of subtasks. The subtask is (Action, Ending). Action is one of [`move', `climb']. The ending is one of [`facing intermediation', `across intermediation', and `to the goal']. Replace the intermediation with the answer to question 1."
    \item \textbf{Perception} ``Where is the \textit{intermediation}? Answer in [x0,y0,x1,y1] format, don't say anything else."
    \item \textbf{Discriminator} ``Is there any \textit{intermediation}? Just answer yes or no." or ``Is the \textit{task} finished at current state?''
\end{enumerate}
Note after the output of prompt 1, all the \textit{intermediation} in prompts will be replaced by the specific intermediation in answer 1. One example output with a specific terrain can be found in Fig.~\ref{fig:system}

\subsubsection{Sub-task Set and Deployment Details}
\label{subsec:subtask}

The sub-task is defined as an \textit{(Action, Ending)} pair. \textit{Action} is one of [`move', `climb'], \textit{Ending} is one of [`facing intermediation', `across intermediation', and `to the goal'].
In real deployments, we find the VLM does not output irrational sub-tasks ``climb facing intermediation'' and ``climb to the goal'', so only four pairs are used: ``Move facing intermediation'', ``Move across intermediation'', ``Move to the goal'', and ``Climb across intermediation''. The sub-task execution module can access sub-task instructions from VLM, RGB-D images, odometry from VIO SLAM, and output a velocity command $(V_x, V_y, V_{yaw})$ to the PAS control policy. The sub-tasks are executed in a closed loop and double-checked by VLM. While one sub-task is \textit{Not} at the \textbf{End Point} judged by VLM, the system consistently executes it.

\begin{figure}[h]
    \centering
    \includegraphics[width=\linewidth]{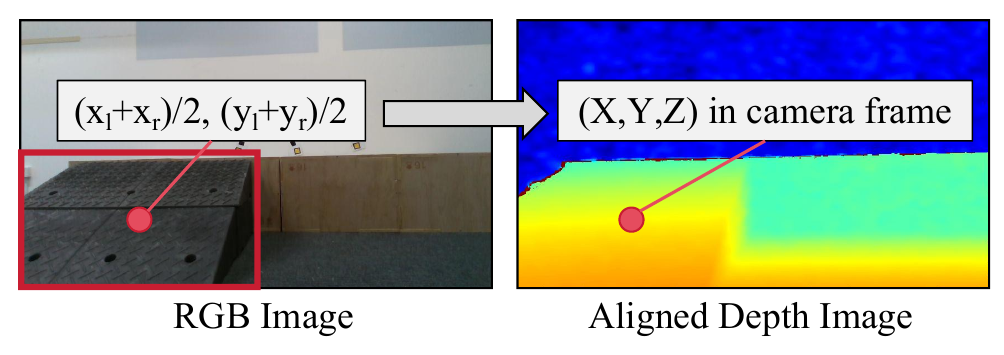}
    \caption{Illustration of the trajectory refinement module combined with depth image.}
    \label{fig:traj_refine}
    \vspace{-3mm}
\end{figure}

Algorithm~\ref{alg:1}: \textbf{``Move facing intermediation''.} The VLM is capable of detecting intermediations and outputting accurate bounding boxes for them. Combining with a depth image input, we can obtain the exact 3D position of the intermediation by $\displaystyle \{X, Y, Z\}=\{\frac{(i-x_{0})\cdot d_{ij}}{f_x},\frac{(j-y_{0})\cdot d_{ij}}{f_y},d_{ij}\}$, and output an accurate sub-goal based on that, as shown in Fig.~\ref{fig:traj_refine}.
    
Algorithm~\ref{alg:2}: \textbf{``Move across intermediation''.} Move forward to the sub-goal located at the same horizontal line as the final goal. 

Algorithm~\ref{alg:3}: \textbf{``Move freely to goal''.} Integrated with the localization module, move freely to the final goal until the distance between the robot and the sub-goal is less than 0.1m.
    
Algorithm~\ref{alg:4}: \textbf{``Climb across intermediation''.} The generalizable locomotion control policy consistently classifies whether the robot is on the plane or is crossing an intermediation, which is discussed in detail in Section \ref{sec:terrain_classify}. If the terrain estimator identifies sudden changes from terrain to plane, the ``done'' signal is thrown by the task execution module.

\begin{algorithm}[h]
  \caption{Move facing intermediation}
  \label{alg:1}
  \begin{algorithmic}
    \Require
        RGB image $C$; Aligned depth image $D$; Odometry $O$
    \Ensure
        Velocity command $cmd_{vel}$;
        \State Query VLM for intermediation bounding box $\{x_l,x_r,y_l,y_r\}$;
        \State Obtain intermediation center in pixel $\{x_0,y_0\}=\{(x_l+x_r)/2, (y_l+y_r)/2\}$;
    \While{\textit{Not} \textbf{$x_0<thereshold$}}
        \State Obtain intermediation center in real 3D position $\displaystyle\{X, Y, Z\}=\{\frac{(i-x_{0})\cdot d_{ij}}{f_x},\frac{(j-y_{0})\cdot d_{ij}}{f_y},d_{ij}\}$;
        \State Transfer sub-goal $\{0,X,0\}$ in robot frame to fixed world frame;
        \While{Distance error greater than 0.1m}
            \State Calculate command velocity $cmd_{vel}$ by PD control based on the sub-goal;
            \State Send $cmd_{vel}$ to PAS control policy;
        \EndWhile
        \State Query VLM for intermediation bounding box $\{x_l,x_r,y_l,y_r\}$;
        \State Obtain intermediation center in pixel $\{x_0,y_0\}=\{(x_l+x_r)/2, (y_l+y_r)/2\}$;
    \EndWhile
  \end{algorithmic}  
\end{algorithm}

\begin{algorithm}[h]
  \caption{Move across intermediation}
  \label{alg:2}
  \begin{algorithmic}
    \Require
        Odometry $O$
    \Ensure
        Velocity command $cmd_{vel}$;
    \While{\textit{Not} \textbf{invisible of the intermediation} judged by VLM discriminator}
        \State Obtain final goal ${X,Y,Z}$ in robot frame;
        \State Transfer sub-goal $\{X,0,0\}$ in robot frame to fixed world frame;
        \While{Distance error greater than 0.1m}
            \State Calculate command velocity $cmd_{vel}$ by PD control based on the sub-goal;
            \State Send $cmd_{vel}$ to PAS control policy;
        \EndWhile
    \EndWhile
  \end{algorithmic}  
\end{algorithm}

\begin{algorithm}[h]
  \caption{Move freely to the goal}
  \label{alg:3}
  \begin{algorithmic}
    \Require
        Odometry $O$
    \Ensure
        Velocity command $cmd_{vel}$;
    \While{\textit{Not} \textbf{finished the task defined by language $\mathcal{L}$} judged by VLM discriminator}
        \State Obtain final goal $\{X,Y,Z\}$;
        \While{Distance error greater than 0.1m}
            \State Calculate command velocity $cmd_{vel}$ by PD control based on the final goal;
            \State Send $cmd_{vel}$ to PAS control policy;
        \EndWhile
    \EndWhile
  \end{algorithmic}  
\end{algorithm}

\begin{algorithm}[H]
  \caption{Climb across intermediation}
  \label{alg:4}
  \begin{algorithmic}
    \Require
        Terrain Classification; Odometry $O$
    \Ensure
        Velocity command $cmd_{vel}$;
    \While{\textit{Not} \textbf{invisible of the intermediation} judged by VLM discriminator}
        \State $Terrain Change Count=0$;
        \While{$Count<2$}
            \State Calculate command velocity $cmd_{vel}$ by PD control to keep moving and facing straight;
            \State Send $cmd_{vel}$ to PAS control policy;
            \State $Count=Count+Classification\oplus Last\_Classification$;
        \EndWhile
    \EndWhile
  \end{algorithmic}  
\end{algorithm}

\subsection{Training Details of Low-level Locomotion Control Policy}

\label{sec:training_detail}

We use the IsaacGym simulator \cite{makoviychuk2021isaac} for policy training and deploy 4,096 quadruped robot agents. The training process has two steps as shown in Fig.~\ref{fig:pipeline}. Both steps use the Proximal Policy Optimization (PPO)\cite{schulman2017proximal} method, and both are trained in 40,000 iterations of exploration and learning. The control policy within the simulator operates at a frequency of 50 Hz.

\subsubsection{Problem Definition}

    We decompose the locomotion control problem into discrete locomotion dynamics. The environment can be fully represented as $\boldsymbol{x}_t$ at each time step $t$, with a discrete time step $d_t=0.02s$.

    \textbf{State Space: }The entire training process includes the following three types of observation information: proprioception $\boldsymbol{p}_t\in\mathbb{R}^{45}$, privileged state $\boldsymbol{s}_t\in\mathbb{R}^{4}$, and terrain information $\boldsymbol{t}_t\in\mathbb{R}^{187}$. Proprioception $\boldsymbol{p}_t\in\mathbb{R}^{45}$ contains gravity vector $\boldsymbol{g}_t^p\in\mathbb{R}^{3}$ and base angular velocity $\boldsymbol{\omega}_t^p\in\mathbb{R}^{3}$ from IMU, velocity command $\boldsymbol{c_t}=\left(v_x^\mathrm{cmd},v_y^\mathrm{cmd},\omega_z^\mathrm{cmd}\right)\in\mathbb{R}^3$, joint positions $\boldsymbol{\theta}_t^p\in\mathbb{R}^{12}$, joint velocities $\boldsymbol{\theta'}_t^p\in\mathbb{R}^{12}$, last action $\boldsymbol{a}_{t-1}^p\in\mathbb{R}^{12}$. Privileged state $\boldsymbol{s}_t\in\mathbb{R}^{4}$ contains base linear velocity $\boldsymbol{v}_t^p\in\mathbb{R}^{3}$ and the ground friction $\boldsymbol{\mu}_t^p\in\mathbb{R}^{1}$. Note that although the base linear velocity can be obtained by integrating the acceleration data from IMU, it has significant errors and accumulates errors over time, hence it cannot be used in the real deployment. Terrain information contains height measurement $\boldsymbol{i}_t^e\in\mathbb{R}^{187}$, which includes 187 height values sampled from the grid surrounding the robot, refer to the yellow point grid surrounding the robot in Fig.~\ref{fig:pipeline}. We have two training steps as shown in Fig.~\ref{fig:pipeline}. Policy in the first step uses all information $\boldsymbol{p}_t\in\mathbb{R}^{45}$,  $\boldsymbol{s}_t\in\mathbb{R}^{4}$, and $\boldsymbol{t}_t\in\mathbb{R}^{187}$ as observation. In the second step and in the real deployment, the policy uses only proprioception $\boldsymbol{p}_t\in\mathbb{R}^{45}$ as observation.

    \textbf{Action Space: }The policy outputs the target positions of 12 joints as the action space $\boldsymbol{a}_t\in\mathbb{R}^{12}$. During real robot deployment, the expected joint positions are sent to the lower-level joint PD controllers $(K_p=40, K_d=0.5)$ for execution via the ROS (Robot Operating System) platform.

\subsubsection{Reward Function}

The reward function is composed of four components: task reward $r_t^T$, survival reward $r_t^A$, performance reward $r_t^E$, and style reward $r_t^S$. And the total reward is the sum of them $r_t = r_t^T + r_t^A + r_t^E + r_t^S$.
Specifically, the task reward mainly consists of the tracking of linear and angular velocities, formulated as the exponent of the velocity tracking error; the alive reward gives a reward to the robot for each step to encourage it not to fall over; the performance reward includes energy consumption, joint velocity, joint acceleration, and angular velocity stability; the style reward includes the time the feet are off the ground and the balance of the forces on the feet, with the hope that the robot can walk with a more natural gait. The details of each reward function are shown in Table ~\ref{tab:reward}.

\begin{table*}[htbp]
    \centering
    \caption{\label{tab:reward}Reward Function}
    \fontsize{10}{12}\selectfont
    \begin{tabular}{cccc}
        \toprule
        \textbf{Type} & \textbf{Item} & \textbf{Formula} & \textbf{Weight}\\ \midrule
        \multirow{2}{*}{\textbf{Task}} & Lin vel & $\exp\left(-\|\mathbf{v}_{t,xy}^\mathrm{des}-\mathbf{v}_{t,xy}\|_2/0.25\right)$ & $3.0$ \\ 
        & Ang vel & $\exp\left(-\|\omega_{t,z}^\mathrm{des}-\omega_{t,z}\|_2/0.25\right)$ & $1.0$ \\ \midrule
        \textbf{Safety} & Alive & 1 & $1.0$ \\ \midrule
        \multirow{4}{*}{\textbf{Performance}} & Energy & $\|\dot{\mathbf{q}}\|_2\cdot\|\tau\|_2$ & $-1\times10^{-6}$ \\
        & Joint vel & $\|\dot{\mathbf{q}}\|_2$ & $-0.002$ \\
        & Joint acc & $\|\ddot{\mathbf{q}}\|_2$ & $-2\times10^{-6}$ \\
        & Ang vel Stability & $(\|\omega_{t,x}\|_2+\|\omega_{t,y}\|_2)$ & $-0.2$ \\ \midrule
        \multirow{2}{*}{\textbf{Style}} & Feet in air & $\sum_{i=0}^{3}\left(\mathbf{t}_{air,i}-0.3\right)+10\cdot\min\left(0.5-\mathbf{t}_{air,i},0\right)$ & $0.05$ \\ 
        & Balance & $\| F_{feet,0}+F_{feet,2}-F_{feet,1}-F_{feet,3} \|_2$ & $-2\times10^{-5}$ \\ 
        \bottomrule
    \end{tabular}
    \vspace{-5mm}
\end{table*}

\subsubsection{Termination Conditions}

We terminate the episode when the robot base's roll angle (the rotation around the forward axis) exceeds 0.8 rad, the robot base's pitch angle (the rotation around the vertical axis) exceeds 1.0 rad, or the robot's position does not change significantly for over 1 second. If the robot does not trigger any termination conditions within 20 seconds or successfully arrives at the edge of one terrain, we also finish this episode and mark this episode as time out.

\subsubsection{Terrain Curriculum}

Previous work \cite{heess2017emergence} has demonstrated that training quadruped robot on various terrains can result in high generalizability and robustness to different ground surfaces. Due to the instability of reinforcement learning in its early stages, it is challenging for robots to directly acquire locomotion skills on complex terrains. Therefore, we employ and refine the ``terrain curriculum'' approach proposed in \cite{rudin2022learning}. Specifically, we create $80$ different terrains, distributed across an $8\times10$ grid. The terrains are divided into $8$ categories, with each type ranging from easy to difficult, consisting of $10$ variations. Each terrain measures $8$ meters in length and width. The first category consists of ascending stairs, with stair heights uniformly increasing from $0$ to $0.2$ meters, and a fixed stair width of $0.3$ meters, designed for training in climbing stairs continuously. The second category features descending stairs, with stair heights uniformly increasing from $0$ to $0.2$ meters, and a fixed stair width of $0.3$ meters, intended for training in descending stairs continuously. The third category comprises ascending platforms, with platform heights uniformly increasing from $0.16$ to $0.22$ meters, and platform widths varying randomly from $0.8$ to $1.5$ meters, used for training to step up onto higher platforms. The fourth category includes descending platforms, with platform heights uniformly increasing from $0.16$ to $0.22$ meters, and platform widths varying randomly from $0.8$ to $1.5$ meters, for training to jump down from higher platforms. The fifth category is for ascending ramps, with ramp angles uniformly increasing from $0$ to $30$ degrees, aimed at training for climbing up ramps. The sixth category is for descending ramps, with ramp angles uniformly increasing from $0$ to $30$ degrees, aimed at training for descending ramps. The seventh category consists of flat ground with no obstacles, for training in walking on level surfaces. The eighth category is rough terrain, with the addition of Perlin noise with amplitudes uniformly increasing from $0$ to $0.15$ meters, for training on uneven surfaces such as rocky roads.

\subsubsection{Dynamic Randomization}

To enhance the robustness and reduce the gap between the simulation and reality, we have a series of randomizations including the mass, the center of gravity position, the initial joint positions, the motor strength, and the coefficient of friction, all of which are subject to random variation within a preset range. Details are in Table~\ref{tab:random}. 

\begin{table}[htbp]
    \centering
    \caption{\label{tab:random}Dynamic randomization}
    \begin{tabular}{ccc}
        \toprule
        \textbf{Parameters} & \textbf{Range} & \textbf{Unit} \\ \midrule
        Base mass & [0, 3] & $kg$\\
        Mass position of X axis & [-0.2, 0.2] & $m$\\
        Mass position of Y axis & [-0.1, 0.1] & $m$\\
        Mass position of Z axis & [-0.05, 0.05] & $m$\\
        Friction & [0, 2] & - \\
        Initial joint positions & [0.5, 1.5]$~\times$nominal value & $rad$\\
        Initial base velocity & [-1.0, 1.0] (all directions) & $m/s$\\
        Motor strength & [0.9, 1.1]$~\times$nominal value & - \\
        \bottomrule
    \end{tabular}
\end{table}

In addition, the observation information obtained by the robot's sensors is also added with random Gaussian noise to simulate the sensor errors that may occur in a real environment. Details are in Table~\ref{tab:gaussian}.

\begin{table}[htbp]
    \centering
    \fontsize{9}{10}\selectfont
    \vspace{-3mm}
    \caption{\label{tab:gaussian}Gaussian noise}
    \begin{tabular}{ccc}
        \toprule
        \textbf{Observation} & \textbf{Gaussian Noise Amplitude} & \textbf{Unit} \\ \midrule
        Linear velocity & 0.05 & $m/s$\\
        Angular velocity & 0.2 & $rad/s$\\
        Gravity & 0.05 & $m/s^2$\\
        Joint position & 0.01 & $rad$\\
        Joint velocity & 1.5 & $rad/s$\\
        \bottomrule
    \end{tabular}
    \vspace{-6mm}
\end{table}

Furthermore, we randomly change the robot's velocity commands every 5 seconds and apply random external forces to the robot every 9 seconds.

\subsubsection{Terrain Classification}
\label{sec:terrain_classify}
After finishing the training process of the PAS control policy, we freeze all the network weight in the PAS control policy and add head to output a boolean terrain classification. The input of the network is $\boldsymbol{p}_t$, $\boldsymbol{s}_t$, $\boldsymbol{t}_t$, and only to predict robot is on the plane or not (on the intermediation). We use BCE Loss as the loss function.

\subsubsection{Network Architecture}

In the first step of training, the terrain encoder $E_{t}$ and the low-level MLP $E_{low}$ are both multilayer perceptrons (MLPs). In the second step of training, the estimator consists of a recurrent neural network (RNN) and a multilayer perceptron (MLP), with the type of recurrent neural network being a Long Short-Term Memory network (LSTM). The specific details of the network are shown in Table ~\ref{tab:network}.

\begin{table}[htbp]
    \caption{\label{tab:network}Details of the network architecture}
    \centering
    \begin{tabular}{cccc}
        \toprule
        \textbf{Network} & \textbf{Input} & \textbf{Hidden layers} & \textbf{Output}\\ \midrule
        $E_{t}$(MLP) & $\boldsymbol{t}_t$ & [128, 64] & $\boldsymbol{t}_{l_t}$\\
        $E_{low}$(MLP) & $\boldsymbol{p}_t$, $\boldsymbol{s}_t$, $\boldsymbol{t}_t$ & [512, 256, 128] & $\boldsymbol{a}_t$\\
        Estimator LSTM & $\boldsymbol{p}_t$ & [256, 256] & $\boldsymbol{h}_t$\\
        Estimator MLP & $\boldsymbol{h}_t$ & [256, 128] & $\boldsymbol{p}_t$\\
        Critic(MLP) & $\boldsymbol{p}_t$, $\boldsymbol{s}_t$, $\boldsymbol{t}_t$ & [512, 256, 128] & $\boldsymbol{V}_t$\\
        Terrain Estimator (MLP) & $\boldsymbol{p}_t$, $\boldsymbol{s}_t$, $\boldsymbol{t}_t$ & [256, 128] & $\boldsymbol{c}_t$\\
        \bottomrule
    \end{tabular}
    \vspace{-2mm}
\end{table}

\subsubsection{Hyperparameters}

The hyperparameters of the PPO algorithm are shown in the Table.~\ref{tab:ppo}:

\begin{table}[htbp]
    \fontsize{10}{12}\selectfont
    \caption{\label{tab:ppo}PPO Hyperparameters}
    \centering
    \begin{tabular}{cc}
        \toprule
        \textbf{Hyperparameter} & \textbf{Value}  \\ \midrule
        clip min std & 0.05 \\
        clip param & 0.2\\
        desired kl & 0.01\\
        entropy coef & 0.01\\
        gamma & 0.99\\
        lam & 0.95\\
        learning rate & 0.001\\
        max grad norm & 1\\
        num mini batch & 4\\
        num steps per env & 24\\
        \bottomrule
    \end{tabular}
    \vspace{-2mm}
\end{table}
\balance
\subsection{Extra Experiments Details of the Low-level Locomotion}

\begin{figure*}[t]
    \centering
    \includegraphics[width=\linewidth]{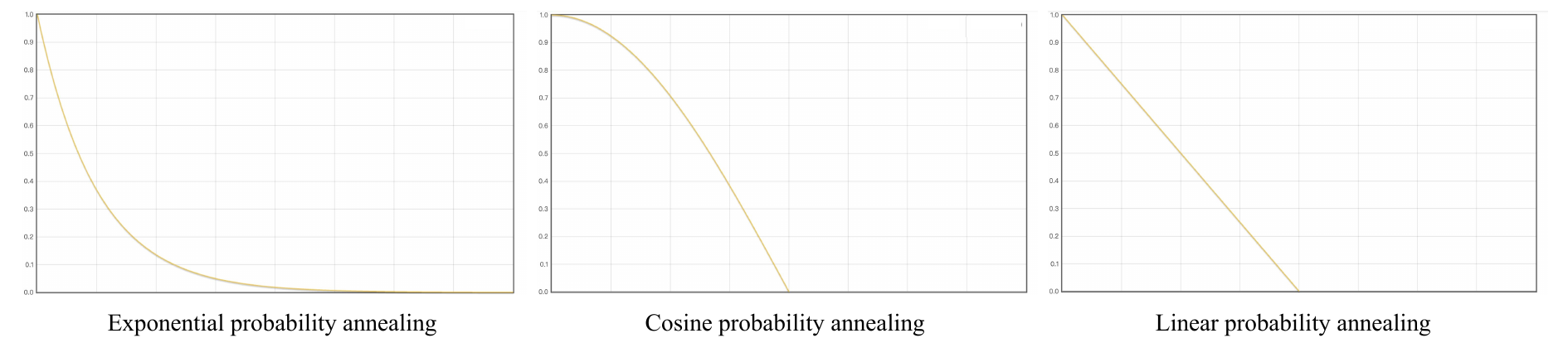}
    \caption{Annealing schedule of different settings. Exponential annealing is fast initially and then slows down, cosine annealing is slow initially and then speeds up, and linear annealing is uniform all the process.}
    \vspace{-5mm}
    \label{fig:schedule}
\end{figure*}

\subsubsection{Metric of Velocity Tracking Ratio}
\begin{align*}
    \text{Linear velocity tracking ratio}&=\exp(-\frac{\|v_{x,y}-v_{x,y}^{\mathrm{target}}\|_2^2}{0.25}),\\\text{Angular velocity tracking ratio}&=\exp(-\frac{\|\omega_{\mathrm{yaw}}-\omega_{\mathrm{yaw}}^{\mathrm{target}}\|_2^2}{0.25}).
\end{align*}

\subsubsection{Comparison Experiments}
\label{sec:comparison}

\begin{itemize}
    \item RMA \cite{kumar2021rma}: A 1D-CNN is used as an adaptation module, employing asynchronously. The teacher-student training framework is used.

    \item IL \cite{wu2023learning}: The first step of training is the same, the second step of training employs the teacher-student framework for imitation learning. The network architectures are the same.

    \item Built-in MPC: The built-in Model Predictive Control (MPC) controller on the Unitree A1 robot (only in physical experiments).
    \vspace{1mm}

    \item Blind: The network architecture is the same as that in the second step of training. Trained only using proprioception directly in one step.

    \item Concurrent \cite{ji2022concurrent}: The policy was trained concurrently with a state estimation network. The training process did not include any input regarding the terrain.

\end{itemize}

\subsubsection{Ablation Experiments}
\label{sec:ablation}

\begin{itemize}
    \item Exp 0.9998: The selection probability decreases exponentially, with a base of 0.9998.

    \item Exp 0.9995: The selection probability decreases exponentially, with a base of 0.9995.

    \item No anneal: The selection probability is set to zero from the beginning, and the predicted hidden state values are used exclusively.

    \item Cosine: The selection probability decreases in the shape of the cosine function on the interval $[0,\pi]$. The rate of probability decrease is initially slow and then accelerates.

    \item Linear: The selection probability decreases in a linear function. The probability decreases uniformly.

\end{itemize}
Specifically, the annealing schedule of exponent, cosine, and linear is shown in Fig.~\ref{fig:schedule}.

\newpage
\bibliographystyle{./IEEEtran} 
\balance
\bibliography{./IEEEabrv,./IEEEexample}

\end{document}